\documentclass[conference]{IEEEtran}

\usepackage[utf8]{inputenc} 
\usepackage[T1]{fontenc}
\usepackage{url}
\usepackage{ifthen}
\usepackage{cite}
\usepackage[cmex10]{amsmath} % Use the [cmex10] option to ensure complicance
\usepackage{amssymb}
\usepackage{mathtools}
\usepackage{bm}
\usepackage{color}
\usepackage{algorithm}
\usepackage{algpseudocode}

\newtheorem{exam}{Example}
\newtheorem{defi}{Definition}
\newtheorem{prop}{Proposition}
\newtheorem{theo}{Theorem}
\newtheorem{lemm}{Lemma}
\newtheorem{condi}{Condition}
\newtheorem{rema}{Remark}
\newtheorem{coro}{Corollary}

\begin{document}
\title{Probability Distribution on Full Rooted Trees} 

% %%% Single author, or several authors with same affiliation:
% \author{%
%   \IEEEauthorblockN{Stefan M.~Moser}
%   \IEEEauthorblockA{ETH Zurich\\
%                     ISI (D-ITET)\\
%                     CH-8092 Zurich, Switzerland\\
%                     Email: moser@isi.ee.ethz.ch}
% }

%%% Several authors with up to three affiliations:
%\author{%
%  \IEEEauthorblockN{Stefan M.~Moser}
%  \IEEEauthorblockA{ETH Zurich\\
%                    ISI (D-ITET), ETH Zentrum\\
%                    CH-8092 Zurich, Switzerland\\
%                    Email: moser@isi.ee.ethz.ch}
%  \and
%  \IEEEauthorblockN{Albus Dumbledore and Harry Potter}
%  \IEEEauthorblockA{Hogwarts School of Witchcraft and Wizardry\\
%                    Hogwarts Castle\\ 
%                    1714 Hogsmeade, Scotland\\
%                    Email: \{dumbledore, potter\}@hogwarts.edu}
%}

%%% Many authors with many affiliations:
 \author{%
   \IEEEauthorblockN{Yuta Nakahara\IEEEauthorrefmark{1},
                     Shota Saito\IEEEauthorrefmark{2},
                     Akira Kamatsuka\IEEEauthorrefmark{3},
                     and Toshiyasu Matsushima\IEEEauthorrefmark{4}}
   \IEEEauthorblockA{\IEEEauthorrefmark{1}%
                     Center for Data Science, Waseda University, 
                     Tokyo, 169-8050, Japan, 
                     yuta.nakahara@aoni.waseda.jp}
   \IEEEauthorblockA{\IEEEauthorrefmark{2}%
                     Faculty of Informatics, Gunma University,
                     Gunma 371-8510, Japan
                     shota.s@gunma-u.ac.jp}
   \IEEEauthorblockA{\IEEEauthorrefmark{3}%
                     Dept. of Information Science, Shonan Institute of Technology,
                     Kanagawa, 251-8511, Japan
                     kamatsuka@info.shonan-it.ac.jp}
   \IEEEauthorblockA{\IEEEauthorrefmark{4}%
                     Dept. of Applied Math., Waseda University,
                     %Department of Applied Mathematics, Waseda University,
                     Tokyo, 169-8555, Japan,
                     toshimat@waseda.jp}
 }

\maketitle

%%%%%%
%% Abstract: 
%% If your paper is eligible for the student paper award, please add
%% the comment "THIS PAPER IS ELIGIBLE FOR THE STUDENT PAPER
%% AWARD." as a first line in the abstract. 
%% For the final version of the accepted paper, please do not forget
%% to remove this comment!
%%
\begin{abstract}
The recursive and hierarchical structure of full rooted trees is applicable to represent statistical models in various areas, such as data compression, image processing, and machine learning. In most of these cases, the full rooted tree is not a random variable; as such, model selection to avoid overfitting becomes problematic. A method to solve this problem is to assume a prior distribution on the full rooted trees. This enables the optimal model selection based on the Bayes decision theory. For example, by assigning a low prior probability to a complex model, the maximum a posteriori estimator prevents the selection of the complex one. Furthermore, we can average all the models weighted by their posteriors. In this paper, we propose a probability distribution on a set of full rooted trees. Its parametric representation is suitable for calculating the properties of our distribution using recursive functions, such as the mode, expectation, and posterior distribution. Although such distributions have been proposed in previous studies, they are only applicable to specific applications. Therefore, we extract their mathematically essential components and derive new generalized methods to calculate the expectation, posterior distribution, etc.
\end{abstract}

%% The paper must be self-contained. However, if you are referring to
%% a full version for checking certain proofs, please provide the
%% publically accessible location below.  If the paper is completely
%% self-contained, you can remove the following line from your
%% submission.
%\textit{A full version of this paper is accessible at:}
%\url{https://arxiv.org/pdf/21xx.xxxx.pdf} 

\section{Introduction}\label{introduction}
\subsection{Objective of this study}
In this paper, we propose a discrete probability distribution on a set with a recursive and hierarchical structure, i.e., a finite set of full rooted trees. Mathematically, a tree is defined as a connected graph without cycles (see, e.g., \cite{graph}). A rooted tree is a tree that has one node known as a root node, and a full rooted tree is a rooted tree in which each inner node has the same number of child nodes. Subsequently, we can define a finite set of subtrees of a full rooted tree. This full rooted tree, which contains all the subtrees in the finite set, is denoted as a base tree herein.

A trivial method to define a probability distribution on this set is to assign occurrence probabilities to all subtrees and regard these values as parameters. In other words, we can define the categorical distribution on the finite set of subtrees of the base tree. However, this definition requires the same number of parameters as the subtrees, which increases in a doubly exponential order of the depth of the base tree.

Therefore, we propose an efficient parametric representation of the probability distribution on a set of subtrees. It is suitable for the recursive structure of full rooted trees and allows the number of parameters to be reduced. Moreover, it enables us to calculate its mode, expectation, posterior distribution, etc., using recursive functions. Therefore, it is efficient from a computational viewpoint. Furthermore, we expect these recursive functions to be effective as a subroutine of the variational Bayesian method and the Markov chain Monte Carlo method in hierarchical Bayesian modeling (see, e.g., \cite{Bishop}).

%Note that the full rooted tree is a completely mathematical object in the above discussion. We are free to represent any kind of applicational object in the real world by the full rooted tree. The set of full rooted trees can be interpreted as various concepts in a similar manner that the support set \{0, 1\} of the Bernoulli distribution can be interpreted as win and loss of a game, front and back of a coin, yes and no of a questionnaire,  etc.

However, our distribution has already been proposed independently in source coding and machine learning, as will be detailed in the next subsection. The novelty of our study is the extraction of the essence from the previous discussion, which depends on the applicational objects, and its representation as a clear mathematical theory. Hence, we derived new generalized recursive algorithms to calculate the expectation, posterior distribution, etc., which could not be derived in previous studies pertaining to real-world applications.

\subsection{Examples of applications}\label{examples_of_application}

Full rooted trees are utilized in various fields of study. For example, for text compression in information theory, a full rooted tree represents a set of contexts, which are strings of the most recent symbols at each time point, and it is known as a context tree\cite{CTW}. In image processing, it represents a variable block-size segmentation, and it is known as quadtree block partitioning\cite{H265}. In machine learning, it represents a nonlinear function that comprises many conditional branches and is known as a decision tree\cite{CART}. In most of these studies, the rooted tree is not a random variable and serves as an index of a statistical model or function, i.e., one full rooted tree $\tau$ corresponds to one statistical model $p(x; \tau)$ or one function $f_\tau (x)$.

Their recursive and hierarchical structures are suitable for representing complex statistical models or functional structures. For example, the expansion of the leaf nodes represents an increase in the contexts of a context tree\cite{CTW}, a division of a block on the image in quadtree partitioning\cite{H265}, and the addition of a conditional branch in the decision tree\cite{CART}. Such expressive capability and extensibility of full rooted trees render them widely applicable in various fields.

However, such hierarchical expressive capability causes a problem in tree selection, i.e., the selection of one statistical model or function. This is because the optimal tree under the criterion of the likelihood or squared loss for training data is inevitably the deepest one. Such a phenomenon is called overfitting in the field of machine learning. Therefore, most previous studies applied a stopping rule for node expansion\cite{H265, CART}, introduced a normalization term into the objective function\cite{XGBoost}, or averaged the statistical models or the functions with some weights\cite{CTW, RF, XGBoost}. However, these algorithmic modifications are heuristic at times.

A theoretical method to solve this problem is to consider the full rooted tree as a random variable and assuming a prior distribution on it. An appropriate prior distribution provides a unified method for selecting one full rooted tree or combining them based on the Bayes decision theory (see, e.g., \cite{Berger}). Although the Bayes decision theory is typically applied to statistical models with unknown continuous parameters, it is also applicable to statistical models with unknown discrete random variables such as full rooted trees (see, e.g., \cite{Bayes_code}). By assigning a high prior probability to a shallow tree and a low prior probability to a deep tree, we can avoid the complex statistical model corresponding to the deep tree.

As mentioned above, most previous studies regard the full rooted tree as a non-stochastic variable. However, few studies adopted the above mentioned approach. In terms of text compression, the complete Bayesian interpretation of context tree weighting method was first investigated by the authors of \cite{CT_th}. Not only the theory, but also the associated algorithm has been improved over the decade it was first investigated (see, e.g., \cite{CT_alg}). Moreover, similar results obtained from rich real data analysis have been reported recently \cite{Papageorgiou, kontoyiannis} (Note that the prior form reported in \cite{Papageorgiou, kontoyiannis} is extremely restricted and cannot be updated as a posterior, in contrast to that reported in \cite{CT_th, CT_alg}). In image processing, the author of \cite{nakahara_entropy} were the first to regard the quadtree as a stochastic model, and its optimal estimation was derived under the Bayes criteria. In machine learning, the authors of \cite{meta-tree} redefined the decision tree as a stochastic generative model and improved most tree weighting methods (e.g., \cite{RF}).

However, these studies depend on specific data or generative models. This might have been the reason that more than 25 years had passed before the first study\cite{CT_th} pertaining to text compression was applied to image processing\cite{nakahara_entropy} and machine learning\cite{meta-tree}. Therefore, we separated the mathematically essential component of the discussion from the modifiable component based on specific data or the generative model. This theoretically expands the potential application of probability distributions on full rooted trees. Subsequently, we derived new generalized methods to evaluate the characteristics of the probability distribution on full rooted trees, which have not been performed in previous studies. More precisely, only Theorems \ref{sum1} and \ref{mode_alg} and Corollary \ref{full-posterior-path} has been used in previous studies. Meanwhile, the other methods expand the possibility of the applying the probability distribution on full rooted trees.

\subsection{Organization of this paper}
The remainder of this paper is organized as follows: In Section \ref{sec-notations}, we present the notations used herein. In Section \ref{sec-definition}, we define the prior on full rooted trees. In Section \ref{sec-properties}, we describe the algorithms for calculating the properties of the proposed distribution, e.g., a marginal distribution for each node, an efficient calculation of the expectation, mode, and the posterior distribution. In Section \ref{discussion}, we discuss usefulness of our distribution in statistical decision theory and hierarchical Bayesian modeling. In Section \ref{sec-future}, we propose some future work. In Section \ref{sec-conclusion}, we conclude the paper.

\section{Notations used for full rooted trees}\label{sec-notations}

\begin{figure}[tbp]
  \centering
  \includegraphics[width=0.45\textwidth]{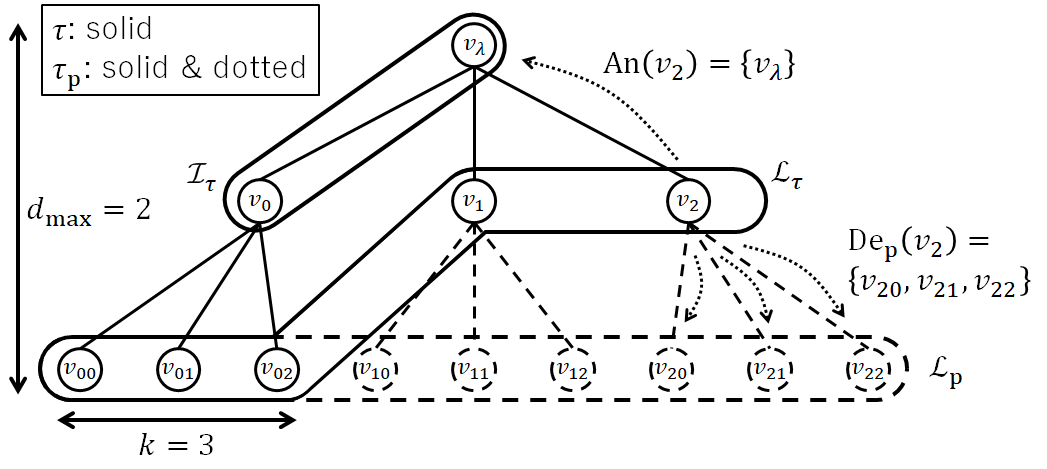}
  % where an .eps filename suffix will be assumed under latex,
  % and a .pdf suffix will be assumed for pdflatex
  \caption{The notations for the rooted trees.}
  \label{notations}
\end{figure}

In this section, we define notations for the rooted trees. They are shown in Fig. \ref{notations}. Let $k \in \mathbb{N}$ denote the maximum number of child nodes and $d_\mathrm{max} \in \mathbb{N}$ denote the maximum depth. Let $\tau_\mathrm{p} = (\mathcal{V}_\mathrm{p}, \mathcal{E}_\mathrm{p})$ denote the perfect\footnote{``Perfect'' means that all inner nodes have exactly $k$ children and all leaf nodes have the same depth.} $k$-ary rooted tree whose depth is $d_\mathrm{max}$ and root node is $v_\lambda$. $\mathcal{V}_\mathrm{p}$ and $\mathcal{E}_\mathrm{p}$ denote the set of the nodes and edges of it, respectively. Then, let $\mathcal{I}_\mathrm{p} \subset \mathcal{V}_\mathrm{p}$, and $\mathcal{L}_\mathrm{p} \subset \mathcal{V}_\mathrm{p}$ denote the set of the inner nodes and the leaf nodes of $\tau_\mathrm{p}$, respectively. For each node $v \in \mathcal{V}_\mathrm{p}$, $\mathrm{Ch}_\mathrm{p}(v) \subset \mathcal{V}_\mathrm{p}$ denote the set of child nodes of $v$ on $\tau_\mathrm{p}$. Notations about the relation between two nodes $v, v' \in \mathcal{V}_\mathrm{p}$ are as follows. Let $v \succ v'$ denote that $v$ is an ancestor node of $v'$, ($v'$ is a descendant node of $v$), $v \succeq v'$ denote that $v$ is an ancestor node of $v'$ or $v'$ itself, ($v'$ is a descendant node of $v$ or $v$ itself), $\mathrm{An}(v) \coloneqq \{ v' \in \mathcal{V}_\mathrm{p} \mid v' \succ v \}$, and $\mathrm{De}_\mathrm{p} (v) \coloneqq \{ v' \in \mathcal{V}_\mathrm{p} \mid v \succ v' \}$.

Subsequently, we consider rooted subtrees of $\tau_\mathrm{p}$ in which their root nodes are the same as $v_\lambda$ and all inner nodes have exactly $k$ children. They are called full rooted subtrees and $\tau_\mathrm{p}$ is called a base tree. Let $\mathcal{T}$ denote the set of all full rooted subtrees of $\tau_\mathrm{p}$. Let $\mathcal{V}_\tau$ and $\mathcal{E}_\tau$ denote the set of the nodes and the edges of $\tau \in \mathcal{T}$, respectively. Let $\mathcal{I}_\tau \subset \mathcal{V}_\tau$, and $\mathcal{L}_\tau \subset \mathcal{V}_\tau$ denote the set of the inner nodes and the leaf nodes of $\tau \in \mathcal{T}$, respectively.

\section{Definition of probability distribution on full rooted subtrees}\label{sec-definition}

In this section, we define a probability distribution on full rooted subtrees $\mathcal{T}$. Let $T$ denote the random variable on $\mathcal{T}$, and $\tau$ denote its realization.

\begin{defi}\label{full-def}
For $(\alpha_{v})_{v \in \mathcal{V}_\mathrm{p}} \in [0, 1]^{|\mathcal{V}_\mathrm{p}|}$, we define probability distribution $p(\tau)$ on $\mathcal{T}$ as below.
\begin{align}
p(\tau) \coloneqq \prod_{v \in \mathcal{I}_\tau} \alpha_v \prod_{v' \in \mathcal{L}_\tau} (1-\alpha_{v'}), \label{definition_of_distribution}
\end{align}
where $\alpha_v = 0$ for $v \in \mathcal{L}_\mathrm{p}$.
\end{defi}

Intuitively, $\alpha_v$ represents the probability that $v$ has child nodes under the condition that $v$ is contained in the tree.\footnote{It will be proved as a theoretical fact in Remark \ref{conditional}.} Therefore, the occurrence probability of a full rooted subtree exponentially decays as its depth increases.

\begin{figure*}[tbp]
  \centering
  \includegraphics[width=0.8\textwidth]{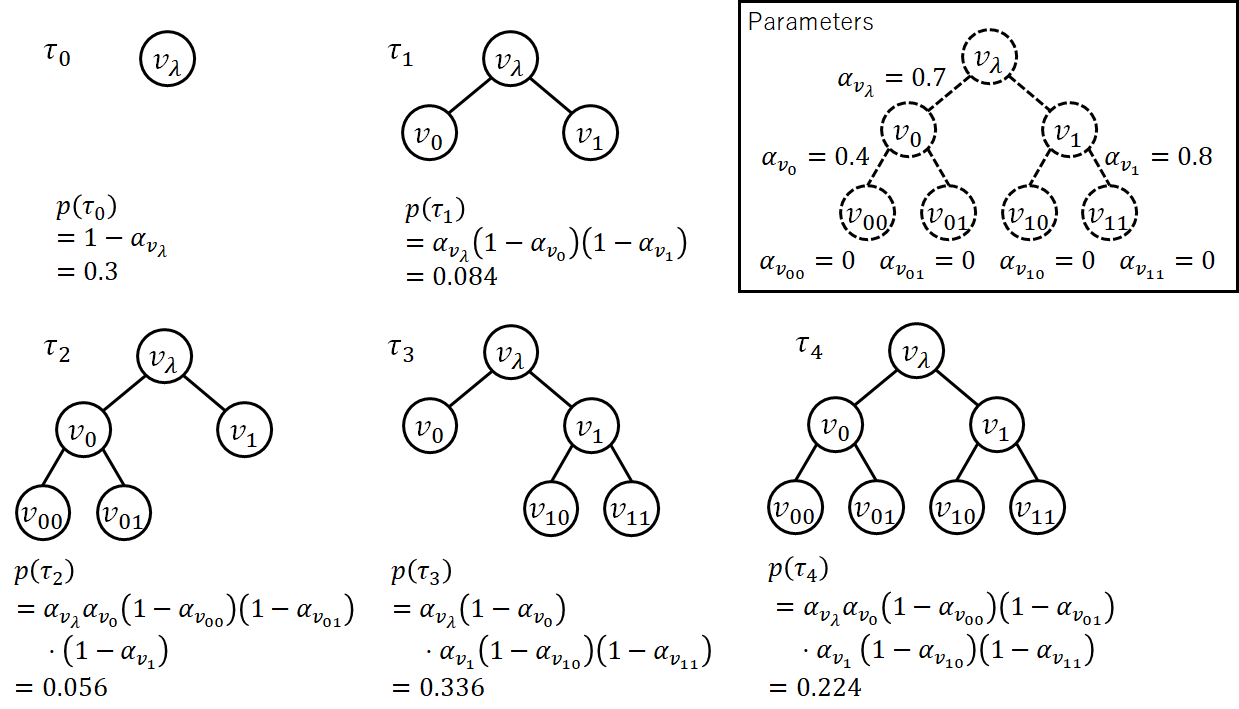}
  % where an .eps filename suffix will be assumed under latex,
  % and a .pdf suffix will be assumed for pdflatex
  \caption{An example of the probability distribution on full rooted subtrees. Here, $k=2$ and $d_\mathrm{max} = 2$. The parameters of the distribution are in the upper right figure. When $k=2$ and $d_\mathrm{max}=2$, $|\mathcal{T}| = 5$. The probability of each full rooted subtree is calculated under the graph of it.}
  \label{example}
\end{figure*}

\begin{exam}
An example of the probability distribution on full rooted subtrees for $k=2$ and $d_\mathrm{max}=2$ is shown in Fig.\ \ref{example}.
\end{exam}

\begin{theo}\label{sum1}
The quantity $p(\tau)$ defined as in (\ref{definition_of_distribution}) fulfills the condition of the probability distribution, that is, $\sum_{\tau \in \mathcal{T}} p(\tau) = 1$.
\end{theo}

\begin{exam}\label{exam2}
Before the proof of Theorem \ref{sum1} for the general case, we describe an example where $d_\mathrm{max} = 2$ and $k = 2$ (see Fig. \ref{example}). First, we factorize the sum as below.
\begin{align}
&\sum_{\tau \in \mathcal{T}} p(\tau) \nonumber \\
&= ( 1-\alpha_{v_\lambda} ) \nonumber \\
&\quad + \alpha_{v_\lambda}(1-\alpha_{v_0})(1-\alpha_{v_1}) \nonumber \\
&\quad + \alpha_{v_\lambda} \alpha_{v_0} (1-\alpha_{v_{00}}) (1-\alpha_{v_{01}}) (1-\alpha_{v_1}) \nonumber \\
&\quad + \alpha_{v_\lambda} (1-\alpha_{v_0}) \alpha_{v_1} (1-\alpha_{v_{10}}) (1-\alpha_{v_{11}}) \nonumber \\
&\quad + \alpha_{v_\lambda} \alpha_{v_0} (1-\alpha_{v_{00}}) (1-\alpha_{v_{01}}) \alpha_{v_1} (1-\alpha_{v_{10}}) (1-\alpha_{v_{11}}) \\
&= ( 1-\alpha_{v_\lambda} ) \nonumber \\
&\quad + \alpha_{v_\lambda} \big\{ (1-\alpha_{v_0})(1-\alpha_{v_1}) \nonumber \\
&\qquad \qquad + \alpha_{v_0} (1-\alpha_{v_{00}}) (1-\alpha_{v_{01}}) (1-\alpha_{v_1}) \nonumber \\
&\qquad \qquad + (1-\alpha_{v_0}) \alpha_{v_1} (1-\alpha_{v_{10}}) (1-\alpha_{v_{11}}) \nonumber \\
&\qquad \qquad + \alpha_{v_0} (1-\alpha_{v_{00}}) (1-\alpha_{v_{01}}) \nonumber \\
&\qquad \qquad \qquad \times \alpha_{v_1} (1-\alpha_{v_{10}}) (1-\alpha_{v_{11}}) \big\} \\
&= ( 1-\alpha_{v_\lambda} ) \nonumber \\
&\quad + \alpha_{v_\lambda} \Big\{ (1-\alpha_{v_0}) \big[ (1-\alpha_{v_1}) + \alpha_{v_1} (1-\alpha_{v_{10}}) (1-\alpha_{v_{11}})\big] \nonumber \\
&\qquad \qquad \quad + \alpha_{v_0} (1-\alpha_{v_{00}}) (1-\alpha_{v_{01}}) \nonumber \\
&\qquad \qquad \qquad \times \big[ (1-\alpha_{v_1}) + \alpha_{v_1} (1-\alpha_{v_{10}}) (1-\alpha_{v_{11}})\big] \Big\} \\
&= ( 1-\alpha_{v_\lambda} ) \nonumber \\
&\quad + \alpha_{v_\lambda} \Big\{ \big[ (1-\alpha_{v_0}) + \alpha_{v_0} (1-\alpha_{v_{00}}) (1-\alpha_{v_{01}}) \big] \nonumber \\
&\qquad \qquad \quad \times \big[ (1-\alpha_{v_1}) + \alpha_{v_1} (1-\alpha_{v_{10}}) (1-\alpha_{v_{11}})\big] \Big\}. \label{exam2_factorized}
\end{align}
Here, $\alpha_{v_{00}} = \alpha_{v_{01}} = \alpha_{v_{10}} = \alpha_{v_{11}} = 0$ since $v_{00}, v_{01}, v_{10}, v_{11} \in \mathcal{L}_\mathrm{p}$. Then,
\begin{align}
(\ref{exam2_factorized}) &= ( 1-\alpha_{v_\lambda} ) \nonumber \\
&\quad + \alpha_{v_\lambda} \Big\{ \big[ (1-\alpha_{v_0}) + \alpha_{v_0} \big] \cdot \big[ (1-\alpha_{v_1}) + \alpha_{v_1} \big] \Big\} \\
&= ( 1-\alpha_{v_\lambda} ) + \alpha_{v_\lambda} \\
&= 1.
\end{align}
\end{exam}

The general proof of Theorem \ref{sum1} is in the following. That also consists of two parts, namely, factorization and substitution. We will first prove Lemma \ref{func_sum}, which is the essential lemma since it is not used only in the proof of Theorem \ref{sum1} but also in the proof of other theorems later.

\begin{lemm}\label{func_sum}
Let $F: \mathcal{T} \to \mathbb{R}$ be a real-valued function on the set $\mathcal{T}$ of the full rooted subtrees of the base tree $\tau_\mathrm{p}$. If $F$ has the form
\begin{align}
F(\tau) = \prod_{v \in \mathcal{I}_\tau} G(v) \prod_{v' \in \mathcal{L}_\tau} H(v'),
\end{align}
where $G: \mathcal{V}_\mathrm{p} \to \mathbb{R}$ and $H: \mathcal{V}_\mathrm{p} \to \mathbb{R}$ are real-valued functions on $\mathcal{V}_\mathrm{p}$, then the summation $\sum_{\tau \in \mathcal{T}} F(\tau)$ can be recursively decomposed as follows.
\begin{align}
\sum_{\tau \in \mathcal{T}} F(\tau) = \phi (v_\lambda),
\end{align}
where $\phi: \mathcal{V}_\mathrm{p} \to \mathbb{R}$ is defined as below.
\begin{align}
\phi (v) \coloneqq
\begin{cases}
H(v), & v \in \mathcal{L}_\mathrm{p},\\
H(v) + G(v) \prod_{v' \in \mathrm{Ch}_\mathrm{p} (v)} \phi (v'), & v \in \mathcal{I}_\mathrm{p}.
\end{cases}\label{phi_v}
\end{align}
\end{lemm}

\textit{Proof:} Let $[v_\lambda]$ denote the tree that consists of only the root node $v_\lambda$ of the base tree $\tau_\mathrm{p}$. Then, the cases of the sum is divided as follows.
\begin{align}
\sum_{\tau \in \mathcal{T}} F(\tau) &= \sum_{\tau \in \mathcal{T}} \prod_{v \in \mathcal{I}_\tau} G(v) \prod_{v' \in \mathcal{L}_\tau} H(v') \label{root_sum} \\
&= \prod_{v \in \mathcal{I}_{[v_\lambda]}} G(v) \prod_{v' \in \mathcal{L}_{[v_\lambda]}} H(v') \nonumber \\
&\qquad + \sum_{\tau \in \mathcal{T} \setminus \{ [v_\lambda] \}} \prod_{v \in \mathcal{I}_\tau} G(v) \prod_{v' \in \mathcal{L}_\tau} H(v')
\\
&= H(v_\lambda) +  \sum_{\tau \in \mathcal{T} \setminus \{ [v_\lambda] \}} \prod_{v \in \mathcal{I}_\tau} G(v) \prod_{v' \in \mathcal{L}_\tau} H(v') \label{division} \\
&= H(v_\lambda) \nonumber \\
&\quad + G(v_\lambda) \sum_{\tau \in \mathcal{T} \setminus \{ [v_\lambda] \}} \prod_{v \in \mathcal{I}_\tau \setminus \{ v_\lambda \}} G(v) \prod_{v' \in \mathcal{L}_\tau} H(v'), \label{factorization}
\end{align}
where (\ref{division}) is because $[v_\lambda]$ has no inner node and its leaf node is only $v_\lambda$; (\ref{factorization}) is because every tree in $\mathcal{T} \setminus [v_\lambda]$ has $v_\lambda$ and the corresponding factor $G(v_\lambda)$.

\begin{figure}[tbp]
  \centering
  \includegraphics[width=0.45\textwidth]{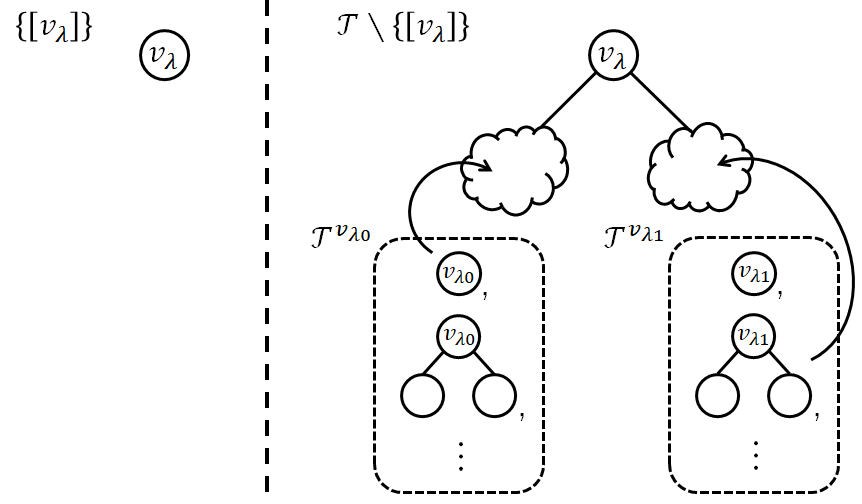}
  % where an .eps filename suffix will be assumed under latex,
  % and a .pdf suffix will be assumed for pdflatex
  \caption{The example of trees in $\{ [v_\lambda] \}$ and $\mathcal{T} \setminus \{ [v_\lambda] \}$, where $k = 2$. The left side shows the structure of the tree in $\{ [v_\lambda] \}$. There is only one tree $[v_\lambda]$. The right side shows the structure of the trees in $\mathcal{T} \setminus \{ [v_\lambda] \}$. All of them have the root node $v_\lambda$ as its inner node. The other structure is determined by choosing subtrees from $\mathcal{T}^{v_{\lambda 0}}$ and $\mathcal{T}^{v_{\lambda 1}}$.}
  \label{factorization_img}
\end{figure}

We have already pointed out that each tree $\tau \in \mathcal{T} \setminus \{ [v_\lambda] \}$ contains $v_\lambda$ as its inner node. The other structure of $\tau$ is determined by the shape of $k$ subtrees whose root nodes are the child nodes of $v_\lambda$ (see Fig. \ref{factorization_img}). We index them in an appropriate order. Then, let $v_{\lambda i}$ denote the $i$-th child node of $v_\lambda$ for $i \in \{ 0, 1, \dots , k-1\}$, i.e. $\{ v_{\lambda 0}, \dots , v_{\lambda \, k-1}\} = \mathrm{Ch}_\mathrm{p} (v_\lambda)$. Let $\mathcal{T}^{v_{\lambda i}}$ denote the set of subtrees whose root node is $v_{\lambda i}$. Then, there is a natural bijection from $\mathcal{T} \setminus \{ [v_\lambda] \}$ to $\mathcal{T}^{v_{\lambda 0}} \times \cdots \times \mathcal{T}^{v_{\lambda \, k-1}}$. Therefore, the summation of (\ref{factorization}) is further factorized. Consequently, we have
\begin{align}
&(\ref{factorization}) = H(v_\lambda) + G(v_\lambda) \nonumber \\
&\times \sum_{(\tau_0, \dots , \tau_{k\!-\!1}) \in \mathcal{T}^{v_{\lambda 0}} \times \cdots \times \mathcal{T}^{v_{\lambda \, k\!-\!1}}} \! \left[ \prod_{v \in \mathcal{I}_{\tau_0}} G(v) \prod_{v' \in \mathcal{L}_{\tau_0}} H(v') \right. \nonumber \\
&\qquad \qquad \qquad \qquad \qquad \quad \left. \cdots \prod_{v \in \mathcal{I}_{\tau_{k\!-\!1}}} G(v) \prod_{v' \in \mathcal{L}_{\tau_{k\!-\!1}}} H(v') \right] \\
&= H(v_\lambda) + G(v_\lambda) \nonumber \\
&\times \sum_{\tau_0 \in \mathcal{T}^{v_{\lambda 0}}} \! \cdots \! \sum_{\tau_{k\!-\!1} \in \mathcal{T}^{v_{\lambda \, k\!-\!1}}} \! \left[ \prod_{v \in \mathcal{I}_{\tau_0}} G(v) \prod_{v' \in \mathcal{L}_{\tau_0}} H(v') \right. \nonumber \\
&\qquad \qquad \qquad \qquad \qquad \quad \left. \cdots \prod_{v \in \mathcal{I}_{\tau_{k\!-\!1}}} G(v) \prod_{v' \in \mathcal{L}_{\tau_{k\!-\!1}}} H(v') \right] \\
&= H(v_\lambda) + G(v_\lambda) \prod_{i=0}^{k-1} \sum_{\tau \in \mathcal{T}^{v_{\lambda i}}} \prod_{v \in \mathcal{I}_{\tau}} G(v) \prod_{v' \in \mathcal{L}_{\tau}} H(v'). \label{child_sum}
\end{align}

Then, from (\ref{root_sum}) and (\ref{child_sum}), we have
\begin{align}
&\underbrace{\sum_{\tau \in \mathcal{T}} \prod_{v \in \mathcal{I}_\tau} G(v) \prod_{v' \in \mathcal{L}_\tau} H(v')}_{(a)} \nonumber \\
&= H(v_\lambda) + G(v_\lambda) \prod_{i=0}^{k-1} \underbrace{\sum_{\tau \in \mathcal{T}^{v_{\lambda i}}} \prod_{v \in \mathcal{I}_{\tau}} G(v) \prod_{v' \in \mathcal{L}_{\tau}} H(v')}_{(b)}. \label{recursive_structure}
\end{align}
The underbraced parts $(a)$ and $(b)$ have the same structure except for the depth of the root node of the subtree. Therefore, $(b)$ can be decomposed in a similar manner from (\ref{root_sum}) to (\ref{child_sum}). We can continue this decomposition to the leaf nodes.

Then, let $\mathcal{T}^{v}$ denote the set of subtrees whose root node is $v \in \mathcal{V}_\mathrm{p}$ in general, i.e., we define a notion similar to $\mathcal{T}^{v_{\lambda i}}$ for not only $v_{\lambda 0}, v_{\lambda 1}, \dots , v_{\lambda \, k-1}$ but also any other nodes $v \in \mathcal{V}_\mathrm{p}$. Finally, we have an alternative definition of $\phi (v): \mathcal{V}_{\mathrm{p}} \to \mathbb{R}$, which is equivalent to (\ref{phi_v}).
\begin{align}
\phi (v) \coloneqq \sum_{\tau \in \mathcal{T}^{v}} \prod_{v' \in \mathcal{I}_{\tau}} G(v') \prod_{v'' \in \mathcal{L}_{\tau}} H(v''),
\end{align}
The equivalence is confirmed by substituting it into both sides of (\ref{recursive_structure}). Therefore, Lemma \ref{func_sum} is proved. \hfill $\blacksquare$

Then, the proof of Theorem \ref{sum1} is as follows.

\textit{Proof:} Using Lemma \ref{func_sum}, we can divide the cases of the sum and factorize the common terms of $\sum_{\tau \in \mathcal{T}} p(\tau)$ in the following recursive manner.
\begin{align}
\sum_{\tau \in \mathcal{T}} p(\tau) = \phi (v_\lambda),
\end{align}
where
\begin{align}
\phi (v) \coloneqq
\begin{cases}
1-\alpha_v, & v \in \mathcal{L}_\mathrm{p},\\
(1-\alpha_v) + \alpha_v \prod_{v' \in \mathrm{Ch}_\mathrm{p} (v)} \phi (v'), & v \in \mathcal{I}_\mathrm{p}. \label{sum_recursive}
\end{cases}\end{align}

Then, we prove $\phi (v) = 1$ for any node $v \in \mathcal{V}_\mathrm{p}$ by structural induction. For any leaf node $v \in \mathcal{L}_\mathrm{p}$, $\alpha_v = 0$ from Definition \ref{full-def}. Therefore,
\begin{align}
\phi (v) = 1-\alpha_v = 1, \qquad v \in \mathcal{L}_\mathrm{p}.
\end{align}

For any inner node $v \in \mathcal{I}_\mathrm{p}$, assuming $\phi (v') = 1$ as the induction hypothesis for any descendant nodes $v' \in \mathrm{De}_\mathrm{p} (v)$,
\begin{align}
\phi (v) &= (1-\alpha_v) + \alpha_v \prod_{v' \in \mathrm{Ch}_\mathrm{p} (v)} \phi (v')\\
&= (1-\alpha_v) + \alpha_v \prod_{v' \in \mathrm{Ch}_\mathrm{p} (v)} 1 \\
&= 1.
\end{align}

Therefore, $\sum_{\tau \in \mathcal{T}} p(\tau) = \phi (v_\lambda) = 1$ since $v_\lambda$ is also in $\mathcal{V}_\mathrm{p}$.

\hfill $\blacksquare$

\begin{rema}
Although Theorem \ref{sum1} is also proved in \cite{nakahara_entropy} and \cite{meta-tree}, we extract the essential part of them as Lemma \ref{func_sum}. In \cite{Papageorgiou, kontoyiannis}, a restricted case of Theorem \ref{sum1} is proved, in which $\alpha_v$ has a common value for all $v \in \mathcal{I}_\mathrm{p}$.
\end{rema}

\section{Properties of probability distribution on full rooted subtrees}\label{sec-properties}

In this section, we describe properties of the probability distribution on full rooted subtrees and methods to calculate them. All the proofs are in Appendix \ref{appendix}. Note that the motivation and usefulness of Conditions \ref{full-product}, \ref{full-sum}, \ref{full-generative-general}, and \ref{full-generative-path} in this section will be described in Section \ref{discussion}.

\subsection{Probability of events on nodes}
At the beginning, we explain why $v \in \mathcal{V}_T$ determines a probabilistic event. We consider any $v \in \mathcal{V}_\mathrm{p}$ is given as a non-stochastic constant and fixed. After that, a full rooted subtree is randomly chosen according to the probability distribution proposed in Section \ref{sec-definition}. Then, $\mathcal{V}_T$ sometimes contains $v$ and sometimes not, depending on the realization $\tau$ of random variable $T$. Therefore, $v \in \mathcal{V}_T$ determines a probabilistic event on $p(\tau)$. Although the probability of such events are trivially represented as $\sum_{\tau \in \mathcal{T}} I \{ v \in \mathcal{V}_\tau \} p(\tau)$, where $I \{ \cdot \}$ denotes the indicator function, we derive computationally efficient forms without the summation about $\tau$ in the following.

\begin{theo}\label{any_node}
For any $v \in \mathcal{V}_\mathrm{p}$, we have the following:
\begin{align}
&\mathrm{Pr} \{ v \in \mathcal{V}_T \} = \prod_{v' \in \mathrm{An}(v)} \alpha_{v'}, \label{any_node_eq}\\
&\mathrm{Pr} \{ v \in \mathcal{I}_T \} = \alpha_v \prod_{v' \in \mathrm{An}(v)} \alpha_{v'}, \label{inner_node_eq}\\
&\mathrm{Pr} \{ v \in \mathcal{L}_T \} = (1-\alpha_v) \prod_{v' \in \mathrm{An}(v)} \alpha_{v'}. \label{leaf_node_eq}
\end{align}
\end{theo}

\begin{exam}
Let us consider $p(\tau)$ shown in Fig. \ref{example}. Trivially, $\mathrm{Pr} \{ v_{01} \in \mathcal{V}_T \}$, $\mathrm{Pr} \{ v_1 \in \mathcal{I}_T \}$, and $\mathrm{Pr} \{ v_0 \in \mathcal{L}_T \}$ are calculated as
\begin{align}
&\mathrm{Pr} \{ v_{01} \in \mathcal{V}_T \} = p(\tau_2) + p(\tau_4) = 0.28, \\
&\mathrm{Pr} \{ v_1 \in \mathcal{I}_T \} = p(\tau_3) + p(\tau_4) = 0.56, \\
&\mathrm{Pr} \{ v_0 \in \mathcal{L}_T \} = p(\tau_1) + p(\tau_3) = 0.42.
\end{align}
The same probabilities are also given by
\begin{align}
&\mathrm{Pr} \{ v_{01} \in \mathcal{V}_T \} = \alpha_{v_\lambda} \alpha_{v_0} = 0.28, \\
&\mathrm{Pr} \{ v_1 \in \mathcal{I}_T \} = \alpha_{v_\lambda} \alpha_{v_1} = 0.56, \\
&\mathrm{Pr} \{ v_0 \in \mathcal{L}_T \} = \alpha_{v_\lambda} (1-\alpha_{v_1}) = 0.42.
\end{align}
\end{exam}

\begin{rema}\label{conditional}
Probabilities of many other events on nodes are derived from Theorem \ref{any_node}. For example,
\begin{align}
\mathrm{Pr} \{ v \in \mathcal{I}_T \mid v \in \mathcal{V}_T \} &= \frac{\mathrm{Pr} \{ v \in \mathcal{I}_T \land v \in \mathcal{V}_T \}}{\mathrm{Pr} \{ v \in \mathcal{V}_T \}} \\
&= \frac{\mathrm{Pr} \{ v \in \mathcal{I}_T \}}{\mathrm{Pr} \{ v \in \mathcal{V}_T \}} \\
&= \alpha_v.
\end{align}
\end{rema}

\subsection{Mode}
We describe an algorithm to find the mode of $p(\tau)$ with $O(k^{d_\mathrm{max}+1})$ computational cost.\footnote{$O (\cdot)$ denotes the Big-O notation, i.e., $f(n) = O(g(n))$ means that $^\exists k > 0, ^\exists n_0 > 0, ^\forall n > n_0, |f(n)| \leq k \cdot g(n)$.} Note that, the size of search space $\mathcal{T}$ is of the order of $\Omega \left(2^{k^{d_\mathrm{max}-2}} \right)$ in general.\footnote{$\Omega ( \cdot )$ denote the Big-Omega notation in complexity theory, i.e., $f(n) = \Omega(g(n))$ means that $^\exists k > 0, ^\exists n_0 > 0, ^\forall n > n_0, f(n) \geq k \cdot g(n)$. $|\mathcal{T}| = \Omega \left(2^{k^{d_\mathrm{max}-2}} \right)$ is proved by substituting $G(v) \equiv H(v) \equiv 1$ in Lemma \ref{func_sum}.} First, replacing all the sum in the proof of Lemma \ref{func_sum} for the max, we can derive the following recursive expression of $\max_{\tau \in \mathcal{T}} p(\tau)$.

\begin{prop}
\begin{align}
\max_{\tau \in \mathcal{T}} p(\tau) = \psi (v_\lambda),
\end{align}
where
\begin{align}
\psi (v) \coloneqq
\begin{cases}
1-\alpha_v = 1, & v \in \mathcal{L}_\mathrm{p},\\
\max \left\{ 1-\alpha_v, \alpha_v \prod_{v' \in \mathrm{Ch}_\mathrm{p}(v)} \psi (v') \right\}, & v \in \mathcal{I}_\mathrm{p}.
\end{cases}
\end{align}
\end{prop}

\begin{exam}
On $p(\tau)$ shown in Fig. \ref{example}, the maximum probability is $p(\tau_3) = 0.336$. It is also calculated as follows.
\begin{align}
&\max \{ 1-\alpha_{v_\lambda}, \alpha_{v_\lambda} \max \{1-\alpha_{v_0}, \alpha_{v_0} \} \nonumber \\
&\qquad \qquad \qquad \quad \times \max \{1-\alpha_{v_1}, \alpha_{v_1}\} \}\\
&= \max \{ 0.3, 0.7 \max \{0.6, 0.4 \} \max \{0.2, 0.8\} \}\\
&= \max \{ 0.3, 0.336 \} = 0.336.
\end{align}
\end{exam}

In addition, we define a flag variable $\delta_v \in \{ 0, 1\}$ as follows.
\begin{defi}
For any $v \in \mathcal{V}_\mathrm{p}$, we define
\begin{align}
\delta_v \coloneqq
\begin{cases}
1, & 1-\alpha_v < \alpha_v \prod_{v' \in \mathrm{Ch}_\mathrm{p} (v)} \psi (v'), \\
0, & \mathrm{otherwise}.
\end{cases}
\end{align}
\end{defi}

We can calculate $\psi(v)$ and $\delta_v$ simultaneously. Then, the mode of $p(\tau)$ is given by the following proposition.

\begin{prop}
$\mathrm{arg} \max_{\tau \in \mathcal{T}} p(\tau)$ is identified as the tree that satisfies
\begin{align}
v \in \mathcal{I}_\tau \Rightarrow \delta_v = 1,\\
v \in \mathcal{L}_\tau \Rightarrow \delta_v = 0.
\end{align}
\end{prop}

Then, the following theorem holds.
\begin{theo}\label{mode_alg}
The mode of $p(\tau)$ can be found by backtracking search from $v_\lambda$ after the calculation of $\psi (v)$ and $\delta_v$. It is detailed in Algorithm \ref{calc_mode} in Appendix \ref{pseudocode}.
\end{theo}

\begin{rema}
In \cite{Papageorgiou, kontoyiannis}, Papageorgiou et al. proposed the same algorithm as Algorithm \ref{calc_mode} as well as an algorithm to find multiple most likely trees on the background of text compression.
\end{rema}

\begin{figure*}[tbp]
  \centering
  \includegraphics[width=0.8\textwidth]{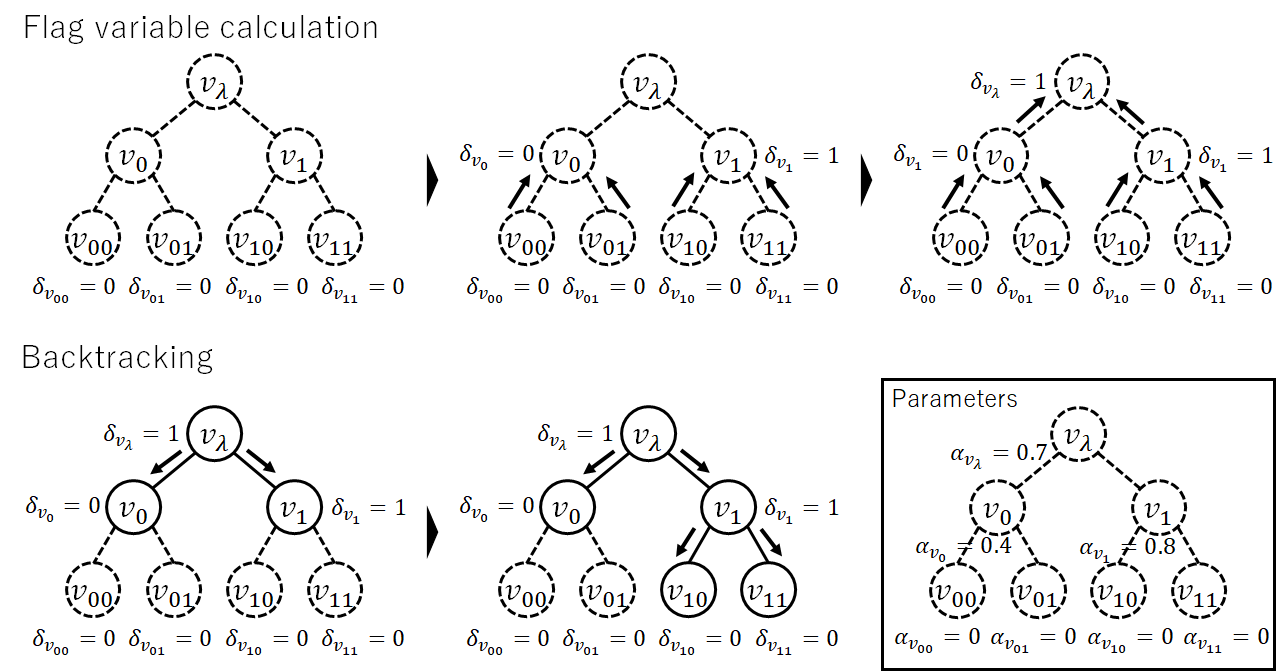}
  % where an .eps filename suffix will be assumed under latex,
  % and a .pdf suffix will be assumed for pdflatex
  \caption{An example of the mode calculation. The parameters are the same as those in Fig. \ref{example} and shown in the lower right figure. Figures on the upper side show the process of the calculation of the flag variable $\delta_v$, which is determined from leaf nodes in order. Figures on the lower side show the process of backtracking. If $\delta_v = 1$, expand the edge. If $\delta_v = 0$, stop the expansion.}
  \label{mode_calculation}
\end{figure*}

\begin{exam}
See Fig. \ref{mode_calculation}. The parameters are the same as those in Fig. \ref{example}. The mode $\tau_3$ is found by the proposed algorithm.
\end{exam}

\subsection{Expectation}
Let $f: \mathcal{T} \to \mathbb{R}$ denote a real-valued function on $\mathcal{T}$. Here, we discuss sufficient conditions of $f$, under which the following expectation can be calculated efficiently with $O(k^{d_\mathrm{max}+1})$ cost.
\begin{align}
\mathbb{E} [f(T)] \coloneqq \sum_{\tau \in \mathcal{T}} f(\tau) p(\tau). \label{expectation}
\end{align}
Note that the size of $\mathcal{T}$ is of the order of $\Omega \left(2^{k^{d_\mathrm{max}-2}} \right)$ in general.

\begin{condi}\label{full-product}
There exist $g: \mathcal{V}_\mathrm{p} \to \mathbb{R}$ and $h: \mathcal{V}_\mathrm{p} \to \mathbb{R}$ such that
\begin{align}
f(\tau) = \prod_{v \in \mathcal{I}_\tau} g(v) \prod_{v' \in \mathcal{L}_\tau} h(v'). \label{product_condition}
\end{align}
\end{condi}

\begin{theo}\label{full-product-ex}
Under Condition \ref{full-product}, we define a recursive function $\phi: \mathcal{V}_\mathrm{p} \to \mathbb{R}$ as 
\begin{align}
\phi (v) \coloneqq
\begin{cases}
h(v), & v \in \mathcal{L}_\mathrm{p},\\
(1-\alpha_v) h(v) \\
\quad + \alpha_v g(v) \prod_{v' \in \mathrm{Ch}_\mathrm{p} (v)} \phi (v'), & v \in \mathcal{I}_\mathrm{p}.
\end{cases}
\end{align}
Then, we can calculate $\mathbb{E} [f(T)]$ as $\mathbb{E} [f(T)] = \phi (v_\lambda)$.
\end{theo}

\begin{exam}
Theorem \ref{any_node} can be regarded examples of Theorem \ref{full-product-ex}.
\end{exam}

\begin{condi}\label{full-sum}
There exist $g: \mathcal{V}_\mathrm{p} \to \mathbb{R}$ and $h: \mathcal{V}_\mathrm{p} \to \mathbb{R}$ such that
\begin{align}
f(\tau) = \sum_{v \in \mathcal{I}_\tau} g(v) + \sum_{v' \in \mathcal{L}_\tau} h(v').
\end{align}
\end{condi}

\begin{theo}\label{full-sum-ex}
Under Condition \ref{full-sum}, we define a recursive function $\xi: \mathcal{V}_\mathrm{p} \to \mathbb{R}$ as 
\begin{align}
\xi (v) \coloneqq
\begin{cases}
h(v), & v \in \mathcal{L}_\mathrm{p},\\
(1-\alpha_v) h(v) \\
\quad + \alpha_v \left(g(v) + \sum_{v' \in \mathrm{Ch}_\mathrm{p} (v)} \xi (v') \right), & v \in \mathcal{I}_\mathrm{p}.
\end{cases}\label{full-sum-ex-recursion}
\end{align}
Then, we can calculate $\mathbb{E} [f(T)]$ as $\mathbb{E} [f(T)] = \xi (v_\lambda)$.
\end{theo}

\begin{rema}
Theorem \ref{full-sum-ex} is useful to calculate the Shannon entropy of $p(\tau)$. It is described in Section \ref{entropy}.
\end{rema}

\subsection{Shannon entropy}\label{entropy}

\begin{coro}\label{theo_entropy}
Substituting $g(v) = - \log \alpha_v$ and $h(v) = - \log (1-\alpha_v)$ into (\ref{full-sum-ex-recursion}), the Shannon entropy $H[T] \coloneqq - \sum_{\tau \in \mathcal{T}}p(\tau) \log p(\tau)$ can be recursively calculated as follows.
\begin{align}
H[T] = \xi (v_\lambda),
\end{align}
where
\begin{align}
\xi (v) \coloneqq
\begin{cases}
0, & v \in \mathcal{L}_\mathrm{p},\\
-(1-\alpha_v) \log (1-\alpha_v) \\
\quad + \alpha_v \left( - \log \alpha_v + \sum_{v' \in \mathrm{Ch}_\mathrm{p} (v)} \xi (v') \right), & v \in \mathcal{I}_\mathrm{p}.
\end{cases}
\end{align}
\end{coro}

\begin{rema}
Kullback-Leibler divergence between two tree distributions $p(\tau)$ and $p'(\tau)$ can be calculated in a similar manner to Corollary \ref{theo_entropy}. This fact may be useful for variational Bayesian inference, in which the Kullback-Leibler divergence is minimized. This is a future work.
\end{rema}

\subsection{Conjugate prior of $p(\tau | \bm \alpha)$}

Here, we consider that $\alpha_v \in [0, 1]$ is also a realization of a random variable. Let $\bm \alpha$ denote $\{ \alpha_v \}_{v \in \mathcal{V}_\mathrm{p}}$, and we describe $p(\tau)$ as $p(\tau | \bm \alpha)$ to emphasize the dependency of $\bm \alpha$ in the following theorem. Then, a conjugate prior for $p(\tau | \bm \alpha)$ is as follows.

\begin{theo}\label{conjugate_prior}
The following probability distribution is a conjugate prior for $p(\tau | \bm \alpha)$.
\begin{align}
p(\bm \alpha) \coloneqq \prod_{v \in \mathcal{V}_\mathrm{p}} \mathrm{Beta} (\alpha_v | \beta_v, \gamma_v),
\end{align}
where $\mathrm{Beta} (\cdot | \beta_v, \gamma_v)$ denotes the probability density function of the beta distribution whose parameters are $\beta_v$ and $\gamma_v$. More precisely,
\begin{align}
p(\bm \alpha | \tau) = \prod_{v \in \mathcal{V}_\mathrm{p}} \mathrm{Beta}(\alpha_v | \beta_{v|\tau}, \gamma_{v|\tau}),
\end{align}
where
\begin{align}
\beta_{v|\tau} &\coloneqq
\begin{cases}
\beta_v + 1, & v \in \mathcal{I}_\tau, \\
\beta_v, & \mathrm{otherwise},
\end{cases}\\
\gamma_{v|\tau} &\coloneqq
\begin{cases}
\gamma_v + 1, & v \in \mathcal{L}_\tau, \\
\gamma_v, & \mathrm{otherwise}.
\end{cases}
\end{align}
\end{theo}

\subsection{$p(\tau)$ as conjugate prior}
We define another random variable $X$ on a set $\mathcal{X}$ and assume $X$ depends on $T$, i.e., it follows a distribution $p(x | \tau)$. Here, we discuss a sufficient condition of $p(x | \tau)$, under which $p(\tau)$ becomes a conjugate prior for it and we can efficiently calculate the posterior $p(\tau | x)$.

\begin{condi}\label{full-generative-general}
There exist two functions $g: \mathcal{V}_\mathrm{p} \times \mathcal{X} \to \mathbb{R}$ and $h: \mathcal{V}_\mathrm{p} \times \mathcal{X} \to \mathbb{R}$, and $p(x | \tau)$ has the following form.
\begin{align}
p(x | \tau) = \prod_{v \in \mathcal{I}_\tau} g(x, v) \prod_{v' \in \mathcal{L}_\tau} h(x, v'). \label{likelihood}
\end{align}
Note that $g$ and $h$ are not necessarily probability density functions.
\end{condi}

\begin{exam}
For given $\mu_1, \mu_2 \in \mathbb{R}$ and $\sigma_1, \sigma_2 \in \mathbb{R}_{>0}$, let $\mathcal{N}(x | \mu_1, \sigma_1^2)$ and $\mathcal{N}(x | \mu_2, \sigma_2^2)$ denote the probability density functions of the normal distributions governed by them. Let $\bm x \coloneqq (x_v)_{v \in \mathcal{V}_\mathrm{p}}$. If we assume
\begin{align}
&g(\bm x, v) = \mathcal{N}(x_v | \mu_1, \sigma_1^2), \\
&h(\bm x, v) = \prod_{v' \in \mathrm{De}_\mathrm{p}(v) \cup \{ v \}} \mathcal{N}(x_{v'} | \mu_2, \sigma_2^2),
\end{align}
we can construct $p(\bm x | \tau)$ that satisfies Condition \ref{full-generative-general}. In other words, the elements of the $|\mathcal{V}_\mathrm{p}|$ dimensional vector $\bm x$ follows the mixture of two normal distributions and  either of the two is chosen by $\tau$.
\end{exam}

\begin{theo}\label{full_posterior_general}
Under Condition \ref{full-generative-general}, we define $q(x | v)$ and $\alpha_{v|x}$ as follows.
\begin{align}
q(x | v) &\coloneqq
\begin{cases}
h(x, v), & v \in \mathcal{L}_\mathrm{p}, \\
(1-\alpha_v) h(x, v) \\
\quad + \alpha_v g(x, v) \prod_{v' \in \mathrm{Ch}_\mathrm{p} (v)} q(x | v'), & v \in \mathcal{I}_\mathrm{p},
\end{cases}\label{q_x_v} \\
\alpha_{v|x} &\coloneqq
\begin{cases}
\alpha_v, & v \in \mathcal{L}_\mathrm{p}, \\
\frac{\alpha_v g(x, v) \prod_{v' \in \mathrm{Ch}_\mathrm{p}(v)} q(x | v')}{q(x |v)}, & v \in \mathcal{I}_\mathrm{p}.
\end{cases}\label{alpha_up_general}
\end{align}
Note that $\alpha_v = 0$ for $v \in \mathcal{L}_\mathrm{p}$ (see Definition \ref{full-def}). Then, the posterior $p(\tau | x)$ is represented as follows.
\begin{align}
p(\tau | x) = \prod_{v \in \mathcal{I}_\tau} \alpha_{v|x} \prod_{v' \in \mathcal{L}_\tau} (1-\alpha_{v'|x}). \label{posterior_general}
\end{align}
\end{theo}

It should be noted that the calculation of $q(x | v)$ and $\alpha_{v|x}$ requires $O(k^{d_\mathrm{max} + 1})$ cost while it requires $\Omega \left( 2^{k^{d_\mathrm{max}-2}} \right)$ cost in general.

Moreover, if we assume the following condition stronger than Condition \ref{full-generative-general}, we can calculate the posterior $p(\tau | x)$ more efficiently with $O(d_\mathrm{max})$ cost.

\begin{condi}\label{full-generative-path}
In addition to Condition \ref{full-generative-general}, we assume that there exist a path from $v_\lambda$ to a leaf node $v_\mathrm{end} \in \mathcal{L}_\mathrm{p}$ and another function $h': \mathcal{V}_\mathrm{p} \times \mathcal{X} \to \mathbb{R}$, which satisfy
\begin{align}
g(x,v) &\equiv 1, \label{g_path}\\
h(x,v) &\coloneqq h'(x,v)^{I \{ v \succeq v_\mathrm{end} \}}. \label{h_path}
\end{align}
Here, $I \{ \cdot \}$ denotes the indicator function. In other words, only $h(x, v)$ on the path from $v_\lambda$ to $v_\mathrm{end}$ takes a value different of 1.
\end{condi}

\begin{coro}\label{full-posterior-path}
Under Condition \ref{full-generative-path}, $q(x | v)$ and $\alpha_{v|x}$ are calculated as follows, more efficiently than (\ref{q_x_v}) and (\ref{alpha_up_general}).
\begin{align}
q(x | v) &=
\begin{cases}
h'(x, v), & v = v_\mathrm{end},\\
(1-\alpha_v) h'(x, v) + \alpha_v q(x | v_\mathrm{ch}), & v \succ v_\mathrm{end},\\
1, & \mathrm{otherwise},
\end{cases}\label{q_x_v_path} \\
\alpha_{v|x} &=
\begin{cases}
\alpha_v, & v \not\succ v_\mathrm{end},\\
\frac{\alpha_v q(x | v_\mathrm{ch})}{q(x |v)}, & v \succ v_\mathrm{end},
\end{cases}\label{alpha_up_path}
\end{align}
where $v_\mathrm{ch}$ is a child node of $v$ on the path from $v_\lambda$ to $v_\mathrm{end}$. Note that we need not calculate $q(x | v)$ for $v \not\succeq v_\mathrm{end}$ to update the posterior and it costs only $O(d_\mathrm{max})$.
\end{coro}

\begin{rema}\label{rem_path}
Condition \ref{full-generative-path} is effective to represent a generation of sequential data $x_1, x_2, \dots , x_N$, in which there exists a path from root node $v_\lambda$ to a leaf node $v_\mathrm{end}^n \in \mathcal{L}_\mathrm{p}$ for each $n \in \{ 1, 2, \dots , N\}$ ($v_\mathrm{end}^n$ and $v_\mathrm{end}^{n'}$ may different each other for $n \neq n'$). The remarkable previous studies using Corollary \ref{full-posterior-path} are \cite{CT_th, CT_alg, Papageorgiou, kontoyiannis, nakahara_entropy, meta-tree} (In \cite{Papageorgiou, kontoyiannis}, only (\ref{q_x_v_path}) is used but (\ref{alpha_up_path}) is not). In other words, they treat only the case under Condition \ref{full-generative-path}. The other theorems in this paper have potential applications to broader fields of study.
\end{rema}

\section{Discussion}\label{discussion}

In this section, we describe the usefulness of our results in statistical decision theory (see, e.g., \cite{Berger}) and hierarchical Bayesian modeling (see, e.g., \cite{Bishop}). First, our results are useful in model selection and model averaging under the Bayes criterion in statistical decision theory (see, e.g., \cite{Berger}). The proposed probability distribution $p(\tau)$ is a conjugate prior for stochastic models $p(x|\tau)$ satisfying Condition \ref{full-generative-general} as shown in Theorem \ref{full_posterior_general}, and the MAP estimate $\mathrm{arg} \max_\tau p(\tau | x)$ can be efficiently calculated by applying Theorem \ref{mode_alg} to the posterior distribution $p(\tau | x)$ obtained by Theorem \ref{full_posterior_general}. This is the Bayes optimal model selection based on the posterior distribution. Furthermore, we can calculate $\sum_\tau p(x_\mathrm{new} | \tau) p(\tau | x)$, i.e., the weighting of the stochastic models based on the posterior distribution, by using Theorems \ref{full_posterior_general} and \ref{full-product-ex} since the stochastic models $p(x|\tau)$ satisfying Condition \ref{full-generative-general} also satisfy Condition \ref{full-product}. This is model averaging of all possible trees with Bayes optimal weights. This corresponds to the methodologies in which they do not select a single tree but aggregate several trees such as \cite{XGBoost, RF}. It should be noted that the occurrence probability of a deep tree exponentially decays in our proposed probability distribution. Therefore, we can avoid the deep tree, which often corresponds to a complex statistical model as mentioned in Section \ref{introduction}.

Second, one example of the applications derived from our results is hyperparameter learning. As mentioned in Remark \ref{rem_path}, Condition \ref{full-generative-path} has been applied to various stochastic models $p(x | \tau)$ in previous studies \cite{CT_th, CT_alg, Papageorgiou, kontoyiannis, nakahara_entropy, meta-tree}. Conditions \ref{full-product} and \ref{full-generative-general} are more generalized conditions than Condition \ref{full-generative-path} since the stochastic model $p(x|\tau)$ satisfying Condition \ref{full-generative-path} also satisfies Conditions \ref{full-product} and \ref{full-generative-general}. In addition, the logarithm of a function $f(\tau)$ satisfying Conditions \ref{full-product} and \ref{full-generative-general} (as well as a stochastic model $p(x|\tau)$ satisfying Condition \ref{full-generative-path}) satisfies Condition \ref{full-sum}. Therefore, we can calculate $\sum_{\tau \in \mathcal{T}} p(\tau | x) \log p(x | \tau)$ by using Theorems \ref{full_posterior_general} and \ref{full-sum-ex}. In particular, the fact that we can calculate the expectations $\mathbb{E}[p(x | T)] = \sum_{\tau \in \mathcal{T}} p(\tau | x) p(x | \tau)$ and $\mathbb{E}[\log p(x | T)] = \sum_{\tau \in \mathcal{T}} p(\tau | x) \log p(x | \tau)$ of the stochastic model $p(x|\tau)$ satisfying Condition \ref{full-generative-path} implies that we can learn hyperparameters of the stochastic models in \cite{CT_th, CT_alg, Papageorgiou, kontoyiannis, nakahara_entropy, meta-tree} by hierarchical Bayesian modeling with variational Bayesian methods (see, e.g., \cite{Bishop}). To the best of our knowledge, there are no unified studies treating hyperparameter learning for these models.

\section{Future work}\label{sec-future}
Since the present study is a theoretical study, the theorems derived will be applied in future studies. Theorems \ref{sum1} and \ref{mode_alg} and Corollary \ref{full-posterior-path} have been used in previous studies \cite{CT_th, CT_alg, Papageorgiou, kontoyiannis, nakahara_entropy, meta-tree}. Therefore, the other theorems can be applied.

In this study, we did not use approximative algorithms such as the variational Bayes or Markov chain Monte Carlo method. Such algorithms are required for learning hierarchical models that contain the probability distribution on full rooted subtrees. The methods proposed herein may serve as a subroutine. The expansion of our methods to approximative algorithms is another future work.

In this study, the class of trees is restricted to that of full trees, in which every inner node has the same number of child nodes. Hence, another the generalization of the class to that of any rooted tree can be considered in future studies.

\section{Conclusion}\label{sec-conclusion}
In this paper, we discuss the probability distribution on full rooted subtrees. Although such a distribution has been used in many fields of studies, such as information theory\cite{CT_th, CT_alg, Papageorgiou, kontoyiannis}, image processing\cite{nakahara_entropy}, and machine learning\cite{meta-tree}, it depends significantly on the specific applications and data generative models. By contrast, we discussed it theoretically, collectively, and independently from a specific data generative model. Subsequently, we derived new generalized methods to evaluate the characteristics of the probability distribution on full rooted subtrees, which have not been performed in previous studies. The derived methods are efficient for calculating the events on the nodes, the mode, the expectation, the Shannon entropy, and the posterior distribution for full rooted subtrees. Therefore, this study expands the possibility of the applying the probability distribution on full rooted subtrees.

\section*{Acknowledgment}

This work was supported by JSPS KAKENHI Grant Numbers JP17K06446, JP19K04914, and JP19K14989.

\bibliographystyle{IEEEtran}
\bibliography{refs_isit}

\begin{thebibliography}{10}
\providecommand{\url}[1]{#1}
\csname url@samestyle\endcsname
\providecommand{\newblock}{\relax}
\providecommand{\bibinfo}[2]{#2}
\providecommand{\BIBentrySTDinterwordspacing}{\spaceskip=0pt\relax}
\providecommand{\BIBentryALTinterwordstretchfactor}{4}
\providecommand{\BIBentryALTinterwordspacing}{\spaceskip=\fontdimen2\font plus
\BIBentryALTinterwordstretchfactor\fontdimen3\font minus
  \fontdimen4\font\relax}
\providecommand{\BIBforeignlanguage}[2]{{%
\expandafter\ifx\csname l@#1\endcsname\relax
\typeout{** WARNING: IEEEtran.bst: No hyphenation pattern has been}%
\typeout{** loaded for the language `#1'. Using the pattern for}%
\typeout{** the default language instead.}%
\else
\language=\csname l@#1\endcsname
\fi
#2}}
\providecommand{\BIBdecl}{\relax}
\BIBdecl

\bibitem{graph}
R.~Kenneth, \emph{Discrete Mathematics and Its Applications}, 7th~ed.\hskip 1em
  plus 0.5em minus 0.4em\relax McGraw-Hill Science, 2011.

\bibitem{Bishop}
\BIBentryALTinterwordspacing
C.~Bishop, \emph{Pattern Recognition and Machine Learning}.\hskip 1em plus
  0.5em minus 0.4em\relax Springer, January 2006. [Online]. Available:
  \url{https://www.microsoft.com/en-us/research/publication/pattern-recognition-machine-learning/}
\BIBentrySTDinterwordspacing

\bibitem{CTW}
F.~M.~J. {Willems}, Y.~M. {Shtarkov}, and T.~J. {Tjalkens}, ``The context-tree
  weighting method: basic properties,'' \emph{IEEE Transactions on Information
  Theory}, vol.~41, no.~3, pp. 653--664, 1995.

\bibitem{H265}
G.~J. {Sullivan}, J.~{Ohm}, W.~{Han}, and T.~{Wiegand}, ``Overview of the high
  efficiency video coding ({HEVC}) standard,'' \emph{IEEE Transactions on
  Circuits and Systems for Video Technology}, vol.~22, no.~12, pp. 1649--1668,
  Dec 2012.

\bibitem{CART}
L.~Breiman, J.~Friedman, C.~J. Stone, and R.~A. Olshen, \emph{Classification
  and regression trees}.\hskip 1em plus 0.5em minus 0.4em\relax CRC press,
  1984.

\bibitem{XGBoost}
\BIBentryALTinterwordspacing
T.~Chen and C.~Guestrin, ``{XGBoost}: A scalable tree boosting system,'' in
  \emph{Proceedings of the 22nd ACM SIGKDD International Conference on
  Knowledge Discovery and Data Mining}, ser. KDD '16.\hskip 1em plus 0.5em
  minus 0.4em\relax New York, NY, USA: ACM, 2016, pp. 785--794. [Online].
  Available: \url{http://doi.acm.org/10.1145/2939672.2939785}
\BIBentrySTDinterwordspacing

\bibitem{RF}
\BIBentryALTinterwordspacing
L.~Breiman, ``\BIBforeignlanguage{English}{Random forests},''
  \emph{\BIBforeignlanguage{English}{Machine Learning}}, vol.~45, no.~1, pp.
  5--32, 2001. [Online]. Available:
  \url{http://dx.doi.org/10.1023/A%3A1010933404324}
\BIBentrySTDinterwordspacing

\bibitem{Berger}
J.~O. Berger, \emph{Statistical Decision Theory and {Bayesian} Analysis}.\hskip
  1em plus 0.5em minus 0.4em\relax Springer Science \& Business Media, 2013.

\bibitem{Bayes_code}
T.~{Matsushima}, H.~{Inazumi}, and S.~{Hirasawa}, ``A class of distortionless
  codes designed by {Bayes} decision theory,'' \emph{IEEE Transactions on
  Information Theory}, vol.~37, no.~5, pp. 1288--1293, Sep. 1991.

\bibitem{CT_th}
T.~Matsushima and S.~Hirasawa, ``A {Bayes} coding algorithm using context
  tree,'' in \emph{Proceedings of 1994 IEEE International Symposium on
  Information Theory}, 1994, p. 386.

\bibitem{CT_alg}
T.~{Matsushima} and S.~{Hirasawa}, ``Reducing the space complexity of a {Bayes}
  coding algorithm using an expanded context tree,'' in \emph{2009 IEEE
  International Symposium on Information Theory}, June 2009, pp. 719--723.

\bibitem{Papageorgiou}
I.~Papageorgiou, I.~Kontoyiannis, L.~Mertzanis, A.~Panotopoulou, and
  M.~Skoularidou, ``Revisiting context-tree weighting for bayesian inference,''
  in \emph{2021 IEEE International Symposium on Information Theory (ISIT)},
  2021, pp. 2906--2911.

\bibitem{kontoyiannis}
\BIBentryALTinterwordspacing
I.~Kontoyiannis, L.~Mertzanis, A.~Panotopoulou, I.~Papageorgiou, and
  M.~Skoularidou, ``{Bayesian} context trees: Modelling and exact inference for
  discrete time series,'' arXiv, 2020. [Online]. Available:
  \url{https://arxiv.org/abs/2007.14900}
\BIBentrySTDinterwordspacing

\bibitem{nakahara_entropy}
\BIBentryALTinterwordspacing
Y.~Nakahara and T.~Matsushima, ``A stochastic model for block segmentation of
  images based on the quadtree and the {Bayes} code for it,'' \emph{Entropy},
  vol.~23, no.~8, 2021. [Online]. Available:
  \url{https://www.mdpi.com/1099-4300/23/8/991}
\BIBentrySTDinterwordspacing

\bibitem{meta-tree}
\BIBentryALTinterwordspacing
N.~Dobashi, S.~Saito, Y.~Nakahara, and T.~Matsushima, ``Meta-tree random
  forest: Probabilistic data-generative model and {Bayes} optimal prediction,''
  \emph{Entropy}, vol.~23, no.~6, 2021. [Online]. Available:
  \url{https://www.mdpi.com/1099-4300/23/6/768}
\BIBentrySTDinterwordspacing

\end{thebibliography}

\appendices

\section{}\label{appendix}

\subsection{Proof of Theorem \ref{any_node}}

First, we prove (\ref{any_node_eq}). Let $I \{ \cdot \}$ denote the indicator function. Then, $\mathrm{Pr} \{ v \in \mathcal{V}_T\}$ is expressed as
\begin{align}
\mathrm{Pr} \{ v \in \mathcal{V}_T \} = \sum_{\tau \in \mathcal{T}} I\{ v \in \mathcal{V}_\tau \} p(\tau).
\end{align}
Here, $v \in \mathcal{V}_\tau$ is equivalent that all the leaf nodes is not a ancestor node of $v$. Then,
\begin{align}
\mathrm{Pr} \{ v \in \mathcal{V}_T \} &= \sum_{\tau \in \mathcal{T}} \prod_{v' \in \mathcal{L}_\tau} I \{ v' \not\succ v \} p(\tau) \\
&= \sum_{\tau \in \mathcal{T}} \prod_{v' \in \mathcal{I}_\tau} \alpha_{v'} \prod_{v'' \in \mathcal{L}_\tau} I \{ v'' \not\succ v \} (1-\alpha_{v''}).
\end{align}

Therefore, using Lemma \ref{func_sum},
\begin{align}
\mathrm{Pr} \{ v \in \mathcal{V}_T \} = \phi_v (v_\lambda),
\end{align}
where
\begin{align}
\phi_v (v') \coloneqq
\begin{cases}
I \{ v' \not\succ v \} (1-\alpha_{v'}), & v' \in \mathcal{L}_\mathrm{p}, \\
I \{ v' \not\succ v \} (1-\alpha_{v'}) \\
\quad + \alpha_{v'} \prod_{v'' \in \mathrm{Ch}_\mathrm{p} (v')} \phi_v (v''), & v' \in \mathcal{I}_\mathrm{p}.
\end{cases} \label{node_event_recursion}
\end{align}
We further transform this function.

If $v' \not\succ v$, then $I \{ v' \not\succ v \} = 1$ and consequently,
\begin{align}
\phi_v (v') =
\begin{cases}
(1-\alpha_{v'}), & v' \in \mathcal{L}_\mathrm{p}, \\
(1-\alpha_{v'}) + \alpha_{v'} \prod_{v'' \in \mathrm{Ch}_\mathrm{p} (v')} \phi_v (v''), & v' \in \mathcal{I}_\mathrm{p}.
\end{cases}\label{not_on_path}
\end{align}
It has the same form as (\ref{sum_recursive}), and every child node $v'' \in \mathrm{Ch}_\mathrm{p} (v')$ also satisfies $v'' \not\succ v$. Therefore, $\phi_v (v') = 1$ for $v' \not\succ v$.

If $v' \succ v$, $v'$ cannot be in $\mathcal{L}_\mathrm{p}$ and has only one child node in $\mathrm{An}(v) \cup \{ v \}$. Let $v'_\mathrm{ch}$ denote it. Then, $\phi_v (v'') = 1$ for the other child nodes $v'' \in \mathrm{Ch}_\mathrm{p}(v') \setminus \{ v'_\mathrm{ch} \}$. Therefore, (\ref{node_event_recursion}) is represented as follows.
\begin{align}
\phi_v (v') = \alpha_{v'} \phi_v (v'_\mathrm{ch}).
\end{align}

Therefore, expanding $\phi_v (v_\lambda)$,
\begin{align}
\mathrm{Pr} \{ v \in \mathcal{V}_T \} = \prod_{v' \in \mathrm{An}(v)} \alpha_{v'}.
\end{align}

Next, we prove (\ref{inner_node_eq}). It is proved in a similar manner to the proof of (\ref{any_node_eq}) since
\begin{align}
\mathrm{Pr} \{ v \in \mathcal{I}_T \} = \sum_{\tau \in \mathcal{T}} \prod_{v' \in \mathcal{L}_\tau} I \{ v' \not\succeq v \} p(\tau).
\end{align}

Lastly, we prove (\ref{leaf_node_eq}). We have
\begin{align}
\mathrm{Pr} \{ v \in \mathcal{L}_T \} = \mathrm{Pr} \{ v \in \mathcal{V}_T \} - \mathrm{Pr} \{ v \in \mathcal{I}_T \}.
\end{align}
Therefore, (\ref{leaf_node_eq}) follows from (\ref{any_node_eq}) and (\ref{inner_node_eq}). \hfill $\blacksquare$

\subsection{Proof of Theorem \ref{full-product-ex}}

Substituting (\ref{product_condition}) into (\ref{expectation}), $\mathbb{E}[f(T)]$ can be represented as follows.
\begin{align}
\mathbb{E} [f(T)] = \sum_{\tau \in \mathcal{T}} \prod_{v \in \mathcal{I}_\tau} \alpha_v g(v) \prod_{v' \in \mathcal{L}_\tau} (1-\alpha_{v'}) h(v').
\end{align}
Then, using Lemma \ref{func_sum}, Theorem \ref{full-product-ex} straightforwardly follows.

\hfill $\blacksquare$

\subsection{Proof of Theorem \ref{full-sum-ex}}

First, we switch the order of the summation as follows.
\begin{align}
\mathbb{E}[f(T)] &= \sum_{\tau \in \mathcal{T}} p(\tau) \left( \sum_{v \in \mathcal{I}_\tau} g(v) + \sum_{v' \in \mathcal{L}_\tau} h(v') \right) \\
&= \sum_{\tau \in \mathcal{T}} p(\tau) \sum_{v \in \mathcal{V}_\mathrm{p}} I\{ v \in \mathcal{I}_\tau \} g(v) \nonumber \\
& \qquad + \sum_{\tau \in \mathcal{T}} p(\tau) \sum_{v' \in \mathcal{V}_\mathrm{p}} I\{ v' \in \mathcal{L}_\tau \} h(v') \\
&= \sum_{v \in \mathcal{V}_\mathrm{p}} g(v) \sum_{\tau \in \mathcal{T}} p(\tau) I\{ v \in \mathcal{I}_\tau \} \nonumber \\
& \qquad + \sum_{v' \in \mathcal{V}_\mathrm{p}} h(v') \sum_{\tau \in \mathcal{T}} p(\tau) I\{ v' \in \mathcal{L}_\tau \} \\
&= \sum_{v \in \mathcal{V}_\mathrm{p}} g(v) \mathrm{Pr} \{ v \in \mathcal{I}_T \} \nonumber \\
& \qquad + \sum_{v' \in \mathcal{V}_\mathrm{p}} h(v') \mathrm{Pr} \{ v' \in \mathcal{L}_T \} \\
&= \sum_{v \in \mathcal{V}_\mathrm{p}} g(v) \alpha_v \prod_{v' \in \mathrm{An}(v)} \alpha_{v'} \nonumber \\
& \qquad + \sum_{v \in \mathcal{V}_\mathrm{p}} h(v) (1-\alpha_v) \prod_{v' \in \mathrm{An}(v)} \alpha_{v'} \label{using_node_events_theorem}\\
&= \sum_{v \in \mathcal{V}_\mathrm{p}} ( \alpha_v g(v) + (1-\alpha_v) h(v) ) \prod_{v' \in \mathrm{An}(v)} \alpha_{v'}, \label{full_sum_ex_before_recursive}
\end{align}
where (\ref{using_node_events_theorem}) is because of Theorem \ref{any_node}.

Next, we decompose the right-hand side of (\ref{full_sum_ex_before_recursive}) until it has the same form as (\ref{full-sum-ex-recursion}). Since $\mathcal{V}_\mathrm{p} = \{ v_\lambda \} \cup \mathrm{De}_\mathrm{p} (v_\lambda)$,
\begin{align}
(\ref{full_sum_ex_before_recursive}) &= \alpha_{v_\lambda} g(v_\lambda) + (1-\alpha_{v_\lambda}) h(v_\lambda) \nonumber \\
&\quad + \sum_{v \in \mathrm{De}_\mathrm{p}(v_\lambda)} ( \alpha_v g(v) + (1-\alpha_v) h(v) ) \prod_{v' \in \mathrm{An}(v)} \alpha_{v'}. \label{full_sum_ex_lambda_divided}
\end{align}
For any $v \in \mathrm{De}_\mathrm{p}(v_\lambda)$, $\mathrm{An}(v)$ contains $v_\lambda$. Therefore,
\begin{align}
(\ref{full_sum_ex_lambda_divided}) &= (1-\alpha_{v_\lambda}) h(v_\lambda) + \alpha_{v_\lambda} \Biggl( g(v_\lambda) \nonumber \\
&\quad + \sum_{v \in \mathrm{De}_\mathrm{p}(v_\lambda)} ( \alpha_v g(v) + (1-\alpha_v) h(v) ) \prod_{v' : v_\lambda \succ v' \succ v} \alpha_{v'}\Biggr). \label{full_sum_ex_lambda_factorized}
\end{align}

Further, since $\mathrm{De}_\mathrm{p} (v_\lambda) =  \bigcup_{v \in \mathrm{Ch}_\mathrm{p}(v_\lambda)} ( \mathrm{De}_\mathrm{p} (v) \cup \{ v \} )$,
\begin{align}
(\ref{full_sum_ex_lambda_factorized}) &= (1-\alpha_{v_\lambda}) h(v_\lambda) + \alpha_{v_\lambda} \Biggl( g(v_\lambda) \nonumber \\
&\quad + \sum_{v \in \mathrm{Ch}_\mathrm{p} (v_\lambda)} \Biggl[ \sum_{v' \in \mathrm{De}_\mathrm{p}(v) \cup \{ v \}} \Big( \alpha_{v'} g(v') \nonumber \\
&\qquad \qquad \qquad \quad + (1-\alpha_{v'}) h(v') \Big) \prod_{v'' : v \succeq v'' \succ v'} \alpha_{v''} \Biggr] \Biggr). \label{full_sum_ex_decomposed}
\end{align}

Comparing (\ref{full_sum_ex_before_recursive}) and (\ref{full_sum_ex_decomposed}), we have
\begin{align}
&\underbrace{\sum_{v \in \mathcal{V}_\mathrm{p}} ( \alpha_v g(v) + (1-\alpha_v) h(v) ) \prod_{v' \in \mathrm{An}(v)} \alpha_{v'}}_{(a)} \nonumber \\
&= (1-\alpha_{v_\lambda}) h(v_\lambda) + \alpha_{v_\lambda} \Biggl( g(v_\lambda) \nonumber \\
&\quad + \sum_{v \in \mathrm{Ch}_\mathrm{p} (v_\lambda)} \Biggl[ \underbrace{\sum_{v' \in \mathrm{De}_\mathrm{p}(v) \cup \{ v \}} \Big( \alpha_{v'} g(v')}_{(b)} \nonumber \\
&\qquad \qquad \qquad \quad \underbrace{+ (1-\alpha_{v'}) h(v') \Big) \prod_{v'' : v \succeq v'' \succ v'} \alpha_{v''}}_{(b)} \Biggr] \Biggr). \label{full_sum_ex_last}
\end{align}
The underbraced parts $(a)$ and $(b)$ have the same structure. Therefore, $(b)$ can be decomposed in a similar manner from (\ref{full_sum_ex_before_recursive}) to (\ref{full_sum_ex_decomposed}). We can continue this decomposition to the leaf nodes. 

Finally, we have an alternative definition of $\xi (v): \mathcal{V}_p \to \mathbb{R}$, which is equivalent to (\ref{full-sum-ex-recursion}).
\begin{align}
\xi (v) \coloneqq & \sum_{v' \in \mathrm{De}_\mathrm{p}(v) \cup \{ v \}} \Big( \alpha_{v'} g(v') \nonumber \\
&\qquad + (1-\alpha_{v'}) h(v') \Big) \prod_{v'' : v \succeq v'' \succ v'} \alpha_{v''}
\end{align}
The equivalence is confirmed by substituting it into both sides of (\ref{full_sum_ex_last}). Therefore, Theorem \ref{full-sum-ex} is proved. \hfill $\blacksquare$

%\subsection{Proof of Corollary \ref{theo_entropy}}
%
%\begin{IEEEproof}
%Corollary \ref{theo_entropy} follows from Theorem \ref{full-sum-ex}, in which we substitute $g(v) = \log \alpha_v$ and $h(v) = \log (1-\alpha_v)$.
%\end{IEEEproof}

\subsection{Proof of Theorem \ref{conjugate_prior}}

By the Bayes theorem, we have
\begin{align}
p(\bm \alpha | \tau) &\propto p(\tau | \bm \alpha) p(\bm \alpha)\\
&= \prod_{v \in \mathcal{I}_\tau} \alpha_v \prod_{v' \in \mathcal{L}_\tau} (1-\alpha_{v'}) \prod_{v'' \in \mathcal{V}_\mathrm{p}} \mathrm{Beta} (\alpha_v'' | \beta_{v''}, \gamma_{v''})\\
&= \prod_{v \in \mathcal{I}_\tau} \alpha_v \mathrm{Beta} (\alpha_v | \beta_v, \gamma_v) \nonumber \\
&\qquad \times \prod_{v' \in \mathcal{L}_\tau} (1-\alpha_{v'}) \mathrm{Beta} (\alpha_v' | \beta_{v'}, \gamma_{v'}) \nonumber \\
&\qquad \times \prod_{v'' \in \mathcal{V}_\mathrm{p} \setminus \mathcal{V}_\tau} \mathrm{Beta} (\alpha_v'' | \beta_{v''}, \gamma_{v''}) \\
&\propto \prod_{v \in \mathcal{V}_\mathrm{p}} \mathrm{Beta}(\alpha_v | \beta_{v|\tau}, \gamma_{v|\tau}),
\end{align}
where we used the conjugate property between the Bernoulli distribution and the beta distribution for each term and
\begin{align}
\beta_{v|\tau} &\coloneqq
\begin{cases}
\beta_v + 1, & v \in \mathcal{I}_\tau, \\
\beta_v, & \mathrm{otherwise},
\end{cases}\\
\gamma_{v|\tau} &\coloneqq
\begin{cases}
\gamma_v + 1, & v \in \mathcal{L}_\tau, \\
\gamma_v, & \mathrm{otherwise}.
\end{cases}
\end{align}
\hfill $\blacksquare$

\subsection{Proof of Theorem \ref{full_posterior_general}}

We prove (\ref{posterior_general}) from the right-hand side to the left.
\begin{align}
&\prod_{v \in \mathcal{I}_\tau} \alpha_{v|x} \prod_{v' \in \mathcal{L}_\tau} (1-\alpha_{v'|x}) \nonumber \\
&= \prod_{v \in \mathcal{I}_\tau} \alpha_{v|x} \prod_{v' \in \mathcal{L}_\tau \cap \mathcal{L}_\mathrm{p}} (1-\alpha_{v'|x}) \prod_{v'' \in \mathcal{L}_\tau \cap \mathcal{I}_\mathrm{p}} (1-\alpha_{v''|x}). \label{three_products}
\end{align}
In the following, we transform each of the above products in order. First, the first product is transformed by substituting (\ref{alpha_up_general}) as follows.
\begin{align}
\prod_{v \in \mathcal{I}_\tau} \alpha_{v|x} = \prod_{v \in \mathcal{I}_\tau} \frac{\alpha_v g(x, v) \prod_{v' \in \mathrm{Ch}_\mathrm{p}(v)} q(x | v')}{q(x |v)}. \label{I_p}
\end{align}

Next, the second product is transformed as follows.
\begin{align}
\prod_{v \in \mathcal{L}_\tau \cap \mathcal{L}_\mathrm{p}} (1-\alpha_{v|x}) &= \prod_{v \in \mathcal{L}_\tau \cap \mathcal{L}_\mathrm{p}} (1-\alpha_v) \label{L_p_1}\\
&= \prod_{v \in \mathcal{L}_\tau \cap \mathcal{L}_\mathrm{p}} \frac{(1-\alpha_v) h(x,v)}{q(x | v)}, \label{L_p_2}
\end{align}
where (\ref{L_p_1}) is because of (\ref{alpha_up_general}) and (\ref{L_p_2}) is because $q(x | v) = h(x,v)$ for $v \in \mathcal{L}_\mathrm{p}$.

Lastly, the third product is transformed as follows.
\begin{align}
&\prod_{v \in \mathcal{L}_\tau \cap \mathcal{I}_\mathrm{p}} (1-\alpha_{v|x}) \nonumber \\
&= \prod_{v \in \mathcal{L}_\tau \cap \mathcal{I}_\mathrm{p}} \left( 1- \frac{\alpha_v g(x, v) \prod_{v' \in \mathrm{Ch}_\mathrm{p}(v)} q(x | v')}{q(x |v)} \right) \label{L_t_1} \\
&= \prod_{v \in \mathcal{L}_\tau \cap \mathcal{I}_\mathrm{p}} \frac{q(x |v) - \alpha_v g(x, v) \prod_{v' \in \mathrm{Ch}_\mathrm{p}(v)} q(x | v')}{q(x |v)} \\
&= \prod_{v \in \mathcal{L}_\tau \cap \mathcal{I}_\mathrm{p}} \frac{(1-\alpha_v) h(x,v)}{q(x | v)}, \label{L_t_2}
\end{align}
where (\ref{L_t_1}) is because of (\ref{alpha_up_general}) and (\ref{L_t_2}) is because of substitution of (\ref{q_x_v}) into $q(x|v)$ at the numerator.

Therefore, we can combine (\ref{I_p}), (\ref{L_p_2}) and (\ref{L_t_2}). Then,
\begin{align}
(\ref{three_products}) &= \prod_{v \in \mathcal{I}_\tau} \frac{\alpha_v g(x, v) \prod_{v' \in \mathrm{Ch}_\mathrm{p}(v)} q(x | v')}{q(x |v)} \nonumber \\
&\qquad \times \prod_{v' \in \mathcal{L}_\tau} \frac{(1-\alpha_{v'}) h(x,v')}{q(x | v')}. \label{telescope}
\end{align}
Here, (\ref{telescope}) is a telescoping product, i.e., $q(x|v)$ appears at once in each of the denominator and the numerator. Therefore, we can cancel them except for $q(x | v_\lambda)$. Then,
\begin{align}
(\ref{telescope}) &= \frac{1}{q(x | v_\lambda)} \prod_{v \in \mathcal{I}_\tau} \alpha_v g(x,v) \prod_{v' \in \mathcal{L}_\tau} (1-\alpha_{v'}) h(x,v') \\
&= \frac{1}{q(x | v_\lambda)} \left( \prod_{v \in \mathcal{I}_\tau} g(x,v) \prod_{v' \in \mathcal{L}_\tau} h(x,v') \right) \nonumber \\
&\qquad \times \left( \prod_{v \in \mathcal{I}_\tau} \alpha_v \prod_{v' \in \mathcal{L}_\tau} (1-\alpha_{v'}) \right) \\
&= \frac{p(x | \tau) p(\tau)}{q(x | v_\lambda)}, \label{bayes_theorem_form}
\end{align}
where we used (\ref{likelihood}) and Definition \ref{full-def}.

In addition, because of Theorem \ref{full-product-ex},
\begin{align}
q(x | v_\lambda) = \mathbb{E} [p(x | T)] = \sum_{\tau \in \mathcal{T}} p(x | \tau) p(\tau) = p(x). \label{marginal}
\end{align}
Therefore, 
\begin{align}
(\ref{bayes_theorem_form}) = \frac{p(x|\tau)p(\tau)}{p(x)} = p(\tau|x).
\end{align}
Then, Theorem \ref{full_posterior_general} holds. \hfill $\blacksquare$

\subsection{Proof of Corollary \ref{full-posterior-path}}

We will prove only $q(x|v) = 1$ for $v \not\succeq v_\mathrm{end}$. Then, (\ref{q_x_v_path}) and (\ref{alpha_up_path}) are straightforwardly derived by substituting it with (\ref{g_path}) and (\ref{h_path}) into (\ref{q_x_v}) and (\ref{alpha_up_general}).

For $v \not\succeq v_\mathrm{end}$, substituting (\ref{g_path}) and (\ref{h_path}) into (\ref{q_x_v}),
\begin{align}
q(x | v) =
\begin{cases}
1, & v \in \mathcal{L}_\mathrm{p},\\
(1-\alpha_v) + \alpha_v \prod_{v' \in \mathrm{Ch}_\mathrm{p} (v)} q(x | v), & v \in \mathcal{I}_\mathrm{p}.
\end{cases}
\end{align}
Since this has the same form as (\ref{not_on_path}), $q(x|v) = 1$ for $v \not\succeq v_\mathrm{end}$ is derived in a similar manner. \hfill $\blacksquare$

%\newpage
\section{Pseudocode to calculate mode of $p(\tau)$}\label{pseudocode}
\begin{figure}[h]
\begin{algorithm}[H]
\caption{Calculation of mode of $p(\tau)$}
\label{calc_mode}
\begin{algorithmic}[1]
\Require $\{ \alpha_v \}_{v \in \mathcal{V}_\mathrm{p}}$
\Ensure $\tau^* = \mathrm{arg} \max_\tau p(\tau)$
\Function {flag\_calculation}{$v$} \Comment {Subroutine}
  \If {$v \in \mathcal{L}_\mathrm{p}$}
    \State $\delta_v \leftarrow 0$
    \State \Return 1
  \ElsIf {$v \in \mathcal{I}_\mathrm{p}$}
    \State $\mathtt{tmp} \leftarrow \prod_{v' \in \mathrm{Ch}_\mathrm{p}(v)}$ \Call {flag\_calculation}{$v'$}
    \If {$1-\alpha_v < \alpha_v \cdot \mathtt{tmp}$}
      \State $\delta_v \leftarrow 1$
      \State \Return $\alpha_v \cdot \mathtt{tmp}$
    \Else
      \State $\delta_v \leftarrow 0$
      \State \Return $1-\alpha_v$
    \EndIf
  \EndIf
\EndFunction
\State
\Function {backtracking}{$v, \mathcal{V}, \mathcal{E}$} \Comment {Subroutine}
  \If {$\delta_v = 0$}
    \State \Return
  \ElsIf {$\delta_v = 1$}
    \State $\mathcal{V} \leftarrow \mathcal{V} \cup \mathrm{Ch}_\mathrm{p}(v)$
    \State $\mathcal{E} \leftarrow \mathcal{E} \cup \bigcup_{v' \in \mathrm{Ch}_\mathrm{p}(v)} (v, v')$
    \ForAll {$v' \in \mathrm{Ch}_\mathrm{p}(v)$}
      \State \Call {backtracking}{$v', \mathcal{V}, \mathcal{E}$}
    \EndFor
    \State \Return
  \EndIf
\EndFunction
\State
\Procedure {}{} \Comment {The main procedure}
\State \Call {flag\_calculation}{$v_\lambda$}
\State $\mathcal{V} \leftarrow \emptyset$
\State $\mathcal{E} \leftarrow \emptyset$
\State \Call {backtracking}{$v_\lambda, \mathcal{V}, \mathcal{E}$}
\State \Return $\tau^* = (\mathcal{V}, \mathcal{E})$
\EndProcedure
\end{algorithmic}
\end{algorithm}
\end{figure}

\end{document}